\pdfoutput=1
\documentclass[runningheads]{llncs}

\usepackage{accv}

\usepackage{accvabbrv}

\usepackage{graphicx}
\usepackage{booktabs}
\usepackage{multirow}
\usepackage{tabularx,array}
\usepackage{afterpage}
\definecolor{tg}{rgb}{0,0.4,0}
\newcommand{\tablegreen}[1]{{\color{tg}{#1}}}

\usepackage{hyperref}

\begin{document}

\title{RGBD-to-3D Object Mesh Refinement via \\ Depth Matching and Symmetry Propagation}

\titlerunning{RGBD-to-3D Object Mesh Refinement}

\author{Ahyun Seo\inst{1,2} \quad
Minsu Cho\inst{1}}

\authorrunning{A.~Seo \& M.~Cho}
\institute{
$^1$POSTECH \quad $^2$KAIST \\
\url{https://ahyunSeo.github.io/MatchPropMesh}}

\maketitle

\begin{abstract}
Single-view 3D reconstructors often produce plausible meshes that disagree with the input view, especially near depth discontinuities and self-occlusions. We present a lightweight, plug-and-play RGBD-to-3D refinement that improves any RGB-to-3D reconstructor without retraining. Given a depth map, we correct the visible surface by bipartite matching to back-projected depth points, mirror these corrections onto the occluded side across a detected symmetry plane, and propagate them with a smoothness solver. Every stage is closed-form, making the method orders of magnitude faster than optimization-heavy test-time refinement. On GSO and OmniObject3D with five backbones, it yields consistent gains, also with monocular pseudo-depth, benefits more from symmetry on symmetric objects, and compares favorably with prior refinement in accuracy and runtime. It further improves an RGB-D-to-mesh reconstructor and transfers to real captures with noisy sensor depth.
\keywords{Single-view 3D reconstruction \and Test-time refinement \and Depth and symmetry priors}
\end{abstract}

\section{Introduction}
\label{sec:intro}
Single-view 3D object reconstruction recovers a full 3D shape from one RGB or RGB-D image, supporting view synthesis, editing, simulation, AR/VR, and robotic interaction. The problem is ill-posed: much of the object is unobserved, and even visible surfaces are ambiguous under sparse texture, specularities, shadows, and occlusion. Reconstructors therefore balance learned shape priors against evidence from the input, and small errors on the visible surface can grow into large distortions over the full mesh.

Most prior work relies on learned priors and synthesized observations: native 3D generative models~\cite{cheng2023sdfusion,wu2018learning,hui2024make,liu2024one,ren2024xcube,xiang2025structured}, diffusion-based lifting of 2D priors through per-instance optimization~\cite{poole2022dreamfusion,chen2023fantasia3d,shi2023mvdream,wang2023prolificdreamer}, feed-forward triplane and transformer pipelines~\cite{hong2023lrm,TripoSR2024,xu2024instantmesh,boss2025sf3d,huang2025spar3d}, and reconstruction from a few generated novel views~\cite{wang2024crm,liu2023syncdreamer,long2024wonder3d}. RGB-D methods exploit depth for stronger geometry~\cite{wu2023multiview,lionar2023nu,wu2024ipod}, but they complete partial point clouds rather than refine RGB-to-3D meshes.

Three gaps remain. First, reconstructed surfaces often look plausible but disagree with the geometry that the input view supports, most noticeably near depth discontinuities and self-occlusions. Second, test-time refinement either relies on costly per-instance optimization~\cite{kim2024metta,shi2025crisp} or on weak 2D signals such as silhouettes~\cite{leung2022blackREFINE}, which offer little guidance for fine surface corrections. Third, occluded regions receive no direct observation. Symmetry is a well-known structural prior~\cite{speciale2016symmetry0convex,zabrodsky1997using}, but prior work imposes it globally or learns it end-to-end rather than using it to transfer reliable corrections into unseen parts.

We propose a lightweight, plug-and-play RGBD-to-3D refinement framework that operates on the output of \emph{any} RGB-to-3D reconstructor without retraining. At inference, a single depth map supervises the geometry, together with the source-view camera pose and intrinsics, which we assume known as in prior refinement or obtain from an off-the-shelf estimator. The framework has three stages. First, bipartite matching between visible mesh vertices and back-projected depth points corrects the observed surface. Second, we estimate a dominant bilateral symmetry plane and mirror these visible corrections onto the occluded side. Third, a smoothness solver propagates the sparse corrections over the full mesh. Every stage is a single closed-form solve, which makes the refinement orders of magnitude faster than optimization-heavy test-time refinement. Experiments on GSO and OmniObject3D with five reconstruction backbones show consistent improvements, which persist with pseudo-depth from an off-the-shelf monocular estimator, larger benefits from symmetry on symmetric objects, and favorable accuracy and runtime against prior refinement methods. The refinement also improves an RGB-D-to-mesh reconstructor and transfers to real captures with noisy sensor depth.

\section{Related Work}
\label{sec:relwork}

\subsubsection{Single-view 3D object reconstruction.}
Native 3D generative models synthesize shapes in volumetric or surface spaces~\cite{cheng2023sdfusion,wu2018learning,hui2024make,liu2024one,ren2024xcube,xiang2025structured}, and diffusion-based methods lift 2D priors to 3D by per-instance optimization that matches rendered views to an image prior~\cite{deng2023nerdi,chen2023fantasia3d,melas2023realfusion,tang2023make,poole2022dreamfusion,shi2023mvdream,wang2023score,wang2023prolificdreamer}. Feed-forward pipelines reconstruct quickly with triplanes and transformers~\cite{hong2023lrm,TripoSR2024,wei2024meshlrm,xu2024instantmesh,boss2025sf3d,huang2025spar3d}, often combined with Gaussian splatting or sparse 3D backbones~\cite{tang2024lgm,xu2024grm,zhang2024gs,liu2024meshformer}, and another line reconstructs from a few generated novel views~\cite{wang2024crm,liu2023one,liu2023syncdreamer,long2024wonder3d,xu2024instantmesh,li2023instant3d,wei2024meshlrm,liu2024meshformer}. These methods favor plausibility under learned priors, and their accuracy with respect to the input view can still lag. RGB-D-to-3D methods use the input depth~\cite{wu2023multiview,lionar2023nu,wu2024ipod} but typically complete partial point clouds rather than produce full, watertight meshes for downstream use. We instead use a single depth map to refine a mesh, improving view-consistent geometry while keeping the mesh output.

\subsubsection{Test-time refinement for 3D.}
Test-time refinement corrects a reconstruction at inference with cues from the given instance, reducing view-specific errors and domain gaps. MeTTA~\cite{kim2024metta} optimizes geometry, appearance, and pose with generative priors and learnable virtual cameras, CRISP~\cite{shi2025crisp} optimizes object pose and shape from RGB-D, and 3D-TTA~\cite{dastmalchi2025test} denoises corrupted point clouds via latent-space diffusion. Other work refines from limited-angle depth~\cite{kulikajevas2022auto}, with recurrent test-time training~\cite{chen2025ttt3r}, or with triplet-based consistency~\cite{yuan2025test3r}. REFINE~\cite{leung2022blackREFINE} is a plug-and-play post-processor that optimizes a per-instance network for silhouette consistency, but mask supervision offers little guidance for fine corrections, and the per-instance optimization adds notable cost. In contrast, our refinement attaches to \emph{any} RGB-to-3D reconstructor and solves every stage in closed form, using metric depth as a hard anchor instead of iterative optimization.

\subsubsection{Symmetry for 3D.}
Symmetry is a long-standing structural prior for reconstruction, completion, and disambiguation. Classical and variational approaches detect and enforce symmetries or symmetrize observations for stability~\cite{speciale2016symmetry0convex,zabrodsky1997using}. From a single image, prior work estimates a reflection plane to create a virtual second view or constrain depth~\cite{koser2011dense}, or fuses flipped predictions with perceptual objectives~\cite{siddique2022sym3dnet}. Symmetry also supports sketch-based modeling with locally symmetric primitives~\cite{hahnlein2022symmetry0sketch}, and recent transformer-based detectors provide scalable cues for 3D generation~\cite{li2025symmetry0strikes}. Unlike methods that enforce symmetry on the predicted shape or learn it end-to-end, we use an estimated bilateral plane to mirror \emph{depth-guided} displacements from the visible side onto the occluded side.

\section{Proposed Method}
\label{sec:method}

\subsection{Problem Definition}
\label{subsec:problem}
\begin{figure*}[t]
    \centering
    \includegraphics[width=\textwidth]{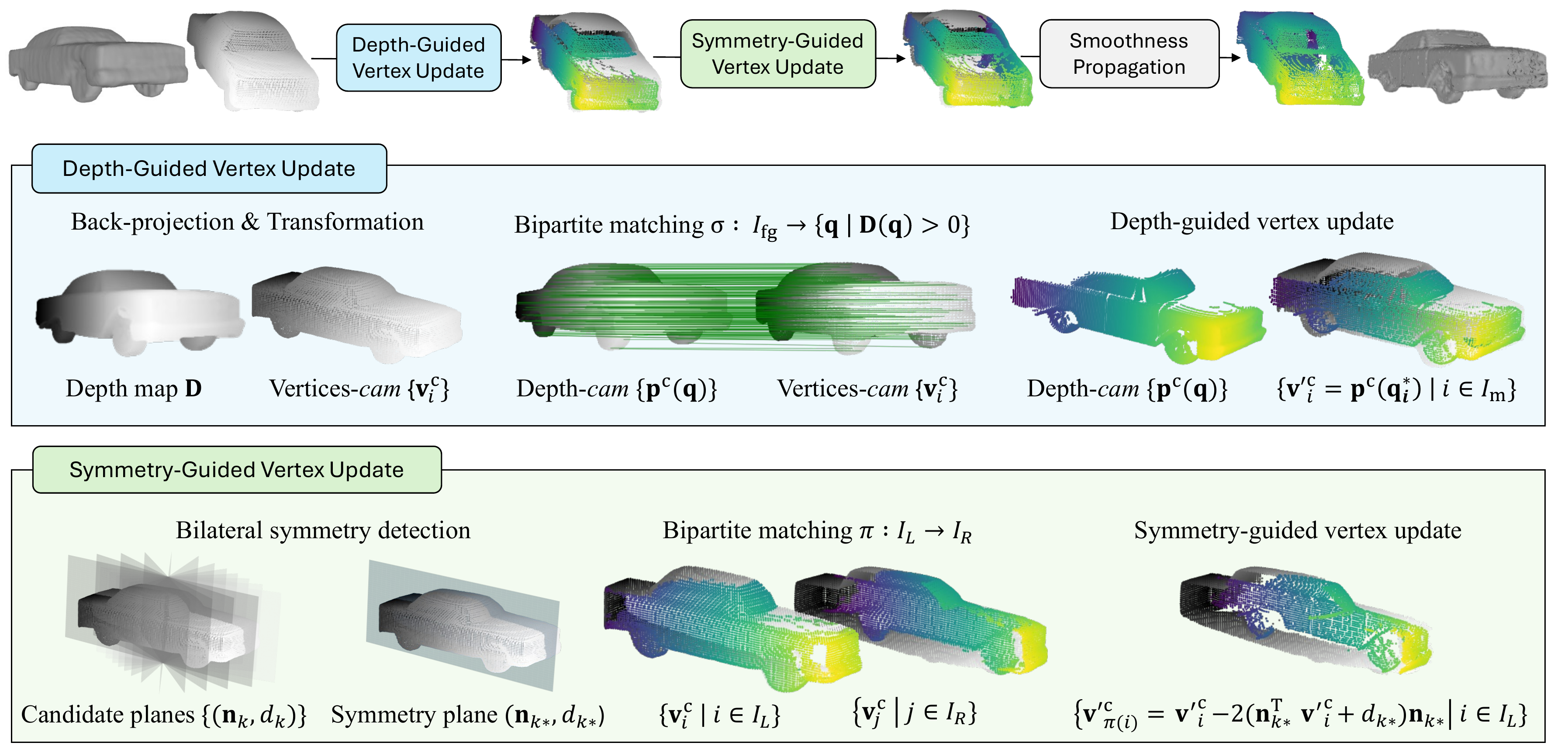}
     \caption{Overview. The depth-guided vertex update rasterizes the camera-frame mesh to find the visible vertices $I_{\mathrm{fg}}$ and moves the matched ones to the back-projected depth points $\mathbf{p}^{\mathrm{c}}(\mathbf{q})$. The symmetry-guided vertex update mirrors these corrections across the detected plane from the camera-facing side ($I_{\mathrm{L}}$) onto the occluded side ($I_{\mathrm{R}}$). A smoothness solver then propagates the sparse displacements over the full mesh.}
    \label{fig:pipeline}
\end{figure*}

Given a single-view RGB image $\mathbf{I}\in\mathbb{R}^{H\times W\times 3}$ and a mesh reconstruction model $\mathcal{S}$, we obtain an initial triangular mesh $(V,F)=\mathcal{S}(\mathbf{I})$ with vertex set $V=\{\mathbf{v}_i\}_{i=1}^{N_v}\subset\mathbb{R}^3$, vertex index set $I=\{1,\dots,N_v\}$, and face set $F=\{f_j\}_{j=1}^{N_f}$ with $f_j=(i_j^1,i_j^2,i_j^3)\in I^3$. Using a metric depth map $\mathbf{D}\in\mathbb{R}^{H\times W}$ of the source view, camera intrinsics $K$, projection matrix $P$, and world-to-camera pose $T_{\mathrm{cw}}$, we estimate per-vertex displacements $\mathbf{d}_i\in\mathbb{R}^3$ that correct the initial mesh toward the observed geometry:
\begin{align}
\mathbf{v}_i' &= \mathbf{v}_i + \mathbf{d}_i,\qquad i\in I,
\end{align}
which yields the updated mesh $(V',F)$ with $V'=\{\mathbf{v}_i'\}_{i\in I}$. As shown in Fig.~\ref{fig:pipeline}, we estimate these displacements in three stages: depth-guided displacement on the visible surface (Sec.~\ref{subsec:depth}), symmetry-guided transfer to the occluded side (Sec.~\ref{subsec:symmetry}), and smoothness propagation over the full mesh (Sec.~\ref{subsec:propagation}).

\subsection{Depth-guided Vertex Displacement}
\label{subsec:depth}
\subsubsection{Motivation.}
Single-view reconstructors are trained for plausible shapes rather than pixel-accurate alignment with the source view, and the visible surface of the initial mesh often deviates from the observed depth. We therefore match visible mesh vertices to back-projected depth points, which yields reliable displacements on the observed surface.

\subsubsection{Camera-frame transformation.}
We express mesh vertices and depth points in the camera frame to compare them directly. A vertex $\mathbf{v}_i$ with homogeneous lift $\tilde{\mathbf{v}}_i$ maps to $\mathbf{v}_i^{\mathrm{c}}=T_{\mathrm{cw}}\tilde{\mathbf{v}}_i$, and an image point $\mathbf{q}=[x,y]^\top$ with depth $\mathbf{D}(\mathbf{q})$ back-projects to
\begin{align}
\mathbf{p}^{\mathrm{c}}(\mathbf{q})={\Big[\tfrac{\mathbf{D}(\mathbf{q})\,(x-c_x)}{f_x},\ \tfrac{\mathbf{D}(\mathbf{q})\,(y-c_y)}{f_y},\ -\mathbf{D}(\mathbf{q})\Big]^{\!\top}},
\end{align}
where $(f_x,f_y,c_x,c_y)$ are the NDC intrinsics and $(x,y)$ the NDC pixel coordinates. We adopt the OpenGL/Blender camera basis ($+X$ right, $+Y$ up, camera looking down $-Z$): points in front of the camera satisfy $z^{\mathrm{c}}<0$ and $\mathbf{D}(\mathbf{q})=-z^{\mathrm{c}}$. The rasterizer receives the same depth attribute $-\mathbf{v}^{\mathrm{c}}_{i,z}$, and mesh and depth points thus share one camera frame.

\subsubsection{Depth guidance for visible vertices.}
Let $\Omega=\{(u,v)\mid \mathbf{D}(u,v)>0\}$ be the foreground pixels of the input depth map. We rasterize the mesh with a triangle rasterizer $\mathcal{R}$ into a rendered depth map $\hat{\mathbf{D}}\in\mathbb{R}^{H\times W}$ and a face map $\mathbf{F}\in\mathbb{R}^{H\times W\times 3}$,
\begin{align}
(\hat{\mathbf{D}},\mathbf{F})=\mathcal{R}(\{\mathbf{v}^{\mathrm{p}}_i\},F,\{-\mathbf{v}_{i,z}^{\mathrm{c}}\};H,W),
\end{align}
where $\mathbf{v}^{\mathrm{p}}_i=P\,\mathbf{v}^{\mathrm{c}}_i$ is the clip-space projection of $\mathbf{v}^{\mathrm{c}}_i$ and $\mathbf{F}$ stores, at each pixel, the three vertex indices of the rendered triangle. The faces rendered at foreground pixels and their vertices form the visible sets
\begin{align}
F_{\mathrm{fg}}=\{\mathrm{set}(\mathbf{F}(u,v))\mid \hat{\mathbf{D}}(u,v)>0\},\qquad I_{\mathrm{fg}}=\textstyle\bigcup F_{\mathrm{fg}} \subseteq I.
\end{align}
We compute a bipartite assignment $\sigma$ between $I_{\mathrm{fg}}$ and $\Omega$ with a GPU auction solver (Sec.~\ref{sec:experiment}) that minimizes the cost $\big\|\mathbf{v}_{i}^{\mathrm{c}}-\mathbf{p}^{\mathrm{c}}(\mathbf{q})\big\|_2$. The assignment is one-to-one but partial: visible vertices typically outnumber the depth points, and we discard matches above a cost threshold, which leaves a matched subset $I_{\mathrm{m}}\subseteq I_{\mathrm{fg}}$ with assigned pixels $\mathbf{q}_i^*=\sigma(i)$ and depth-guided displacements $\mathcal{D}^{\mathrm{c}}=\{\mathbf{d}^{\mathrm{c}}_i\}$,
\begin{align}
\mathbf{d}^{\mathrm{c}}_i=
\begin{cases}
\mathbf{p}^{\mathrm{c}}(\mathbf{q}_i^*)-\mathbf{v}^{\mathrm{c}}_i, & i\in I_{\mathrm{m}},\\
\mathbf{0}, & i \in I \setminus I_{\mathrm{m}}.
\end{cases}
\end{align}

\subsection{Symmetry-guided Vertex Displacement}
\label{subsec:symmetry}
\subsubsection{Motivation.}
Single-view reconstructors trained on large shape collections tend to reproduce the bilateral symmetry of symmetric objects. Depth guidance corrects only the visible surface, which breaks this symmetry and leaves the occluded region unconstrained. As the corrected visible surface is a reliable proxy for its mirror image, we transfer the depth-driven corrections across an estimated symmetry plane into the occluded region.

\subsubsection{Bilateral symmetry plane detection.}
From the initial vertices, we select a dominant bilateral symmetry plane among $N_p$ candidates with azimuth $\theta_k$ and polar angle $\alpha_k$, whose normal (y-up) and offset are
\begin{align}
\mathbf{n}_k=\big(\sin\alpha_k\cos\theta_k,\ \cos\alpha_k,\ \sin\alpha_k\sin\theta_k\big)^\top,\qquad d_k=-\,\mathbf{n}_k^\top \bar{\mathbf{v}}^{\mathrm{c}},
\end{align}
which place the plane through the centroid $\bar{\mathbf{v}}^{\mathrm{c}}$. Each vertex $\mathbf{v}_i^{\mathrm{c}}$ then has a signed distance to the plane and a projection onto it,
\begin{align}
s_i^{(k)}=\mathbf{n}_k^\top \mathbf{v}_i^{\mathrm{c}} + d_k,\qquad \mathbf{p}_i^{(k)}=\mathbf{v}_i^{\mathrm{c}} - s_i^{(k)}\mathbf{n}_k.
\end{align}
Any off-the-shelf detector can provide the plane; we score candidates with a simple rasterized criterion. Instead of reflecting points and accumulating nearest-neighbor distances, we rasterize the signed distances at the in-plane coordinates $(u_i^{(k)},v_i^{(k)})$ of $\mathbf{p}_i^{(k)}$ into two maps $\mathbf{M}^{(k)}_{\pm}\in\mathbb{R}^{H'\times W'}$:
\begin{align}
\mathbf{M}^{(k)}_{+}(u_i^{(k)}, v_i^{(k)}) = s_i^{(k)} \ \text{ if } s_i^{(k)} > 0,\qquad
\mathbf{M}^{(k)}_{-}(u_i^{(k)}, v_i^{(k)}) = s_i^{(k)} \ \text{ if } s_i^{(k)} < 0.
\end{align}
After smoothing both maps with a $5\times5$ mean filter into $\widetilde{\mathbf{M}}^{(k)}_{\pm}$, we score each plane by the average $\ell_1$ discrepancy between the two channels over the cells $\Omega_k$ where both are populated, and select
\begin{align} k^\star &= {\arg\min_{k}\ \frac{1}{|\Omega_k|}\sum_{(u,v)\in\Omega_k}\big| \widetilde{\mathbf{M}}^{(k)}_{+}(u,v) + \widetilde{\mathbf{M}}^{(k)}_{-}(u,v)\big|.} \end{align}
As the negative channel stores signed values, a symmetric cell gives $\widetilde{\mathbf{M}}^{(k)}_{+}+\widetilde{\mathbf{M}}^{(k)}_{-}\approx0$; Fig.~\ref{fig:sd_map} in the supplementary material shows that the discrepancy is minimized at the correct plane ($150^\circ$).

\subsubsection{Displacement transfer by bipartite matching.}
Given the detected plane, we take the depth-matched vertices on its positive side and all vertices on its negative side,
\begin{align}
I_{\mathrm{L}}=\{ i\in I_{\mathrm{m}}\mid s_i^{(k^\star)}>0\},\qquad
I_{\mathrm{R}}=\{ j\in I \mid s_j^{(k^\star)}<0\},
\end{align}
and compute a bipartite assignment $\pi:I_{\mathrm{L}}\!\to\! I_{\mathrm{R}}$ with the same auction solver on in-plane coordinates and unsigned distances,
\begin{align}
C_{ij}=\big\|\boldsymbol{\phi}(i)-\boldsymbol{\phi}(j)\big\|_2,\qquad
\boldsymbol{\phi}(i)=\big[u_i^{(k^\star)}, v_i^{(k^\star)}, |s_i^{(k^\star)}|\big]^\top.
\end{align}
For each matched pair, we move $\pi(i)$ to the reflection of the corrected vertex $\mathbf{v}'^{\mathrm{c}}_i=\mathbf{v}^{\mathrm{c}}_i+\mathbf{d}^{\mathrm{c}}_i$ and set its displacement accordingly:
\begin{align}
\mathbf{v}'^{\mathrm{c}}_{\pi(i)}&=\mathbf{v}'^{\mathrm{c}}_i-2\big(\mathbf{n}_{k^\star}^\top\mathbf{v}'^{\mathrm{c}}_i+d_{k^\star}\big)\,\mathbf{n}_{k^\star},\\
\mathbf{d}^{\mathrm{c}}_{\pi(i)}&=\mathbf{v}'^{\mathrm{c}}_{\pi(i)}-\mathbf{v}^{\mathrm{c}}_{\pi(i)}.
\end{align}
Unlike reflecting $\mathbf{d}^{\mathrm{c}}_i$ alone, this also absorbs the matching residual. We discard matches above a cost threshold and those whose target already carries a depth-driven displacement ($\pi(i)\in I_{\mathrm{m}}$); the reflection therefore never overwrites a depth-driven displacement, and unmatched vertices keep a zero displacement. The camera-facing positive half ($s^{(k^\star)}>0$) serves as the source, and the propagation below fills the occluded vertices on that side.

\subsection{Mesh Refinement with Vertex Displacement}
\label{subsec:propagation}
\subsubsection{Motivation.}
The previous stages provide reliable displacements only on a sparse set of vertices. Applying them directly introduces discontinuities, and leaving the remaining vertices unchanged yields inconsistent geometry. We therefore propagate these displacements as Dirichlet handles with Laplacian smoothing.

\subsubsection{Displacement propagation via Dirichlet handles.}
The handle set $\mathcal{H}=I_{\mathrm{m}}\cup\{\pi(i):i\in I_{\mathrm{L}}\}$ carries the depth-guided displacements of the matched visible vertices and their reflections on the mirrored occluded vertices, written $\{\mathbf{d}^{\mathrm{c}}_i\}_{i\in\mathcal{H}}$. We solve for displacements $\mathbf{u}_i\in\mathbb{R}^3$ of all vertices such that the refined mesh $V'=\{\mathbf{v}^{\mathrm{c}}_i+\mathbf{u}_i\}_{i\in I}$ is smooth and satisfies the handle constraints. With the cotangent Laplacian $L$ of $(V,F)$, we minimize
\begin{align}
E(\mathbf{u})=\tfrac12\sum_{c\in\{x,y,z\}}\mathbf{u}_{\cdot,c}^\top L\,\mathbf{u}_{\cdot,c}\quad \text{s.t.}\quad \mathbf{u}_i=\mathbf{d}^{\mathrm{c}}_i,\ i\in\mathcal{H}.
\label{eq:energy_function}
\end{align}
Splitting the vertices into free indices $\mathcal{F}=I\setminus\mathcal{H}$ and handles $\mathcal{H}$ and partitioning $L$ accordingly gives the reduced linear system $L_{ff}\,\mathbf{u}_{f,c}=-L_{fh}\,\mathbf{d}^{\mathrm{c}}_{h,c}$, which we solve once for all free vertices. For robustness, we clamp the cotangent weights to give degenerate or near-degenerate triangles zero weight in Eq.~\eqref{eq:energy_function}, stabilize $L_{ff}$ with a small Tikhonov term $\tau I$, and softly anchor free boundary vertices with a quadratic penalty $\gamma\|\mathbf{u}_i\|^2$. These terms improve conditioning without visibly changing the result. We update the vertices as $\mathbf{v}_i^{\mathrm{c}\prime}=\mathbf{v}_i^{\mathrm{c}}+\mathbf{u}_i$ and map them back to the world frame, which realizes the displacements $\mathbf{d}_i$ of Sec.~\ref{subsec:problem}. Propagating the depth handles $I_{\mathrm{m}}$ alone yields our visible-only variant, to which we apply a few Taubin smoothing iterations to suppress residual high-frequency artifacts; including the reflected handles yields the full result.

\section{Experiments}
\label{sec:experiment}
\subsection{Datasets and Evaluation}
\subsubsection{Datasets.}
We evaluate on two datasets of real-world scanned objects. Google Scanned Objects (GSO)~\cite{downs2022gso} contains 1{,}030 high-quality textured scans of common household items, all of which we use. OmniObject3D~\cite{wu2023omniobject3d} covers a large vocabulary of about 6{,}000 scanned objects in 190 daily categories; following MeshFormer~\cite{liu2024meshformer}, we take up to five shapes per category (1{,}038 shapes).

\subsubsection{Metrics.}
Following~\cite{xu2024instantmesh,wei2024meshlrm}, we report two geometric metrics, Chamfer Distance (CD; the mean nearest-neighbor distance between the two surfaces, averaged over both directions) and F-score (FS; the harmonic mean of precision and recall), and image metrics on rendered views (SSIM, PSNR, LPIPS, and CLIP similarity).

\subsubsection{Protocol.}
For each shape, we render RGB and depth with BlenderProc~\cite{denninger2023blenderproc2} from 24 viewpoints (12 azimuths $\times$ two elevations $\{15^\circ,30^\circ\}$) and uniformly sample one of them as the input view, fixed across all methods. Unless stated otherwise, we refine with the GT camera pose and the rendered GT depth of this view; \emph{pseudo-depth} denotes the UniDepth-V2~\cite{piccinelli2025unidepthv2} monocular prediction, either rescaled to the GT depth range or affinely fit to the depth rendered from the initial mesh (no GT), and each experiment states its pose and depth sources. We align each mesh candidate to the GT independently, including the initial mesh and the refinement baselines: we transform it into the GT frame, scale its longest side to $0.8$, and resolve the global orientation with a coarse-to-fine rotation search over $\mathrm{SO}(3)$. Our refinement starts from the aligned initial mesh, and we neither re-align nor rescale its output. Before refinement, we decimate meshes above 30k vertices. For the geometric metrics, we sample 16k points from the prediction and the GT and score FS at thresholds $0.02$, $0.05$, and $0.1$; for the image metrics, we compare 48 rendered views with the GT renders.

\begin{table}[t]
\centering
\caption{Mesh refinement results on GSO and OmniObject3D.}
\label{tab:table_main}
\vspace{-2mm}
\resizebox{\textwidth}{!}{%
\setlength{\tabcolsep}{4pt}
\begin{tabular}{llcccccc}
\toprule
& \multirow{2}{*}{Model}
& \multicolumn{2}{c}{Chamfer Distance $\downarrow$}
& \multicolumn{2}{c}{F-score (0.05) $\uparrow$}
& \multicolumn{2}{c}{F-score (0.1) $\uparrow$}\\
\cmidrule(lr){3-4} \cmidrule(lr){5-6} \cmidrule(lr){7-8}
& & Initial & Refined (diff) & Initial & Refined (diff) & Initial & Refined (diff)\\
\midrule
\multirow{5}{*}{\rotatebox{90}{GSO}}
& LGM~\cite{tang2024lgm} & 0.0438 & 0.0306 \tablegreen{{\small(-0.0132)}} & 0.6715 & 0.7983 \tablegreen{{\small(+0.1268)}} & 0.8931 & 0.9320 \tablegreen{{\small(+0.0389)}} \\
& CRM~\cite{wang2024crm} & 0.0365 & 0.0248 \tablegreen{{\small(-0.0117)}} & 0.7473 & 0.8585 \tablegreen{{\small(+0.1112)}} & 0.9346 & 0.9627 \tablegreen{{\small(+0.0281)}} \\
& SF3D~\cite{boss2025sf3d} & 0.0352 & 0.0236 \tablegreen{{\small(-0.0116)}} & 0.7658 & 0.8693 \tablegreen{{\small(+0.1035)}} & 0.9369 & 0.9620 \tablegreen{{\small(+0.0251)}} \\
& SPAR3D~\cite{huang2025spar3d} & 0.0356 & 0.0248 \tablegreen{{\small(-0.0108)}} & 0.7671 & 0.8597 \tablegreen{{\small(+0.0926)}} & 0.9321 & 0.9554 \tablegreen{{\small(+0.0233)}} \\
& InstantMesh~\cite{xu2024instantmesh} & 0.0283 & 0.0205 \tablegreen{{\small(-0.0078)}} & 0.8358 & 0.8950 \tablegreen{{\small(+0.0592)}} & 0.9582 & 0.9721 \tablegreen{{\small(+0.0139)}} \\
\midrule
\multirow{5}{*}{\rotatebox{90}{Omni}}
& LGM~\cite{tang2024lgm} & 0.0397 & 0.0291 \tablegreen{{\small(-0.0106)}} & 0.7185 & 0.8119 \tablegreen{{\small(+0.0934)}} & 0.9031 & 0.9312 \tablegreen{{\small(+0.0281)}} \\
& CRM~\cite{wang2024crm} & 0.0334 & 0.0233 \tablegreen{{\small(-0.0101)}} & 0.7833 & 0.8697 \tablegreen{{\small(+0.0864)}} & 0.9411 & 0.9666 \tablegreen{{\small(+0.0255)}} \\
& SF3D~\cite{boss2025sf3d} & 0.0311 & 0.0210 \tablegreen{{\small(-0.0101)}} & 0.8075 & 0.8910 \tablegreen{{\small(+0.0835)}} & 0.9524 & 0.9701 \tablegreen{{\small(+0.0177)}} \\
& SPAR3D~\cite{huang2025spar3d} & 0.0331 & 0.0238 \tablegreen{{\small(-0.0093)}} & 0.7945 & 0.8673 \tablegreen{{\small(+0.0728)}} & 0.9368 & 0.9552 \tablegreen{{\small(+0.0184)}} \\
& InstantMesh~\cite{xu2024instantmesh} & 0.0320 & 0.0229 \tablegreen{{\small(-0.0091)}} & 0.8011 & 0.8720 \tablegreen{{\small(+0.0709)}} & 0.9383 & 0.9547 \tablegreen{{\small(+0.0164)}} \\
\bottomrule
\end{tabular}
}

\vspace{3mm}
\caption{2D visual metrics of initial and refined meshes.}
\label{tab:table_2d}
\vspace{-2mm}
\resizebox{\textwidth}{!}{%
\setlength{\tabcolsep}{4pt}
\begin{tabular}{llcccccc}
\toprule
& \multirow{2}{*}{Model}
& \multicolumn{2}{c}{SSIM $\uparrow$}
& \multicolumn{2}{c}{PSNR $\uparrow$}
& \multicolumn{2}{c}{LPIPS $\downarrow$}\\
\cmidrule(lr){3-4} \cmidrule(lr){5-6} \cmidrule(lr){7-8}
& & Initial & Refined (diff) & Initial & Refined (diff) & Initial & Refined (diff)\\
\midrule
\multirow{5}{*}{\rotatebox{90}{GSO}}
& LGM~\cite{tang2024lgm} & 0.8148 & 0.8269 \tablegreen{{\small(+0.0121)}} & 14.7640 & 15.5800 \tablegreen{{\small(+0.8160)}} & 0.2892 & 0.2658 \tablegreen{{\small(-0.0234)}} \\
& CRM~\cite{wang2024crm} & 0.8292 & 0.8418 \tablegreen{{\small(+0.0126)}} & 15.9370 & 16.7550 \tablegreen{{\small(+0.8180)}} & 0.2504 & 0.2302 \tablegreen{{\small(-0.0202)}} \\
& SF3D~\cite{boss2025sf3d} & 0.8348 & 0.8422 \tablegreen{{\small(+0.0074)}} & 15.6650 & 16.1650 \tablegreen{{\small(+0.5000)}} & 0.2321 & 0.2219 \tablegreen{{\small(-0.0102)}} \\
& SPAR3D~\cite{huang2025spar3d} & 0.8244 & 0.8314 \tablegreen{{\small(+0.0070)}} & 15.3860 & 15.9430 \tablegreen{{\small(+0.5570)}} & 0.2360 & 0.2255 \tablegreen{{\small(-0.0105)}} \\
& InstantMesh~\cite{xu2024instantmesh} & 0.8413 & 0.8495 \tablegreen{{\small(+0.0082)}} & 17.4310 & 18.0690 \tablegreen{{\small(+0.6380)}} & 0.2078 & 0.1967 \tablegreen{{\small(-0.0111)}} \\
\midrule
\multirow{5}{*}{\rotatebox{90}{Omni}}
& LGM~\cite{tang2024lgm} & 0.8174 & 0.8305 \tablegreen{{\small(+0.0131)}} & 16.1070 & 16.9260 \tablegreen{{\small(+0.8190)}} & 0.2725 & 0.2521 \tablegreen{{\small(-0.0204)}} \\
& CRM~\cite{wang2024crm} & 0.8326 & 0.8466 \tablegreen{{\small(+0.0140)}} & 17.3270 & 18.2820 \tablegreen{{\small(+0.9550)}} & 0.2374 & 0.2143 \tablegreen{{\small(-0.0231)}} \\
& SF3D~\cite{boss2025sf3d} & 0.8342 & 0.8435 \tablegreen{{\small(+0.0093)}} & 15.3700 & 15.9700 \tablegreen{{\small(+0.6000)}} & 0.2238 & 0.2085 \tablegreen{{\small(-0.0153)}} \\
& SPAR3D~\cite{huang2025spar3d} & 0.8287 & 0.8348 \tablegreen{{\small(+0.0061)}} & 15.3730 & 15.7710 \tablegreen{{\small(+0.3980)}} & 0.2252 & 0.2135 \tablegreen{{\small(-0.0117)}} \\
& InstantMesh~\cite{xu2024instantmesh} & 0.8335 & 0.8436 \tablegreen{{\small(+0.0101)}} & 18.1100 & 18.8680 \tablegreen{{\small(+0.7580)}} & 0.2213 & 0.2050 \tablegreen{{\small(-0.0163)}} \\
\bottomrule
\end{tabular}
}

\end{table}

\subsection{Implementation Details}
\subsubsection{Refinement.} We rasterize the mesh with \texttt{nvdiffrast} at the BlenderProc camera (FoV $\approx49^\circ$) at $256\times256$, from $512\times512$ input images, and back-project the input depth at $128\times128$.
For the depth-guided update, we discard matches with an assignment cost above $0.1$.
For the symmetry-guided update, we sweep the plane azimuth over $1^\circ$--$180^\circ$ and the elevation over $\{0^\circ,\dots,30^\circ\}$, and discard left/right matches with a cost above $0.2$.
For propagation, we clamp the cotangent Laplacian weights to $[0,50]$, stabilize the free-free system with small Tikhonov and boundary-anchor terms, and apply 5 Taubin smoothing iterations to the depth-only mesh.
We use these values in all experiments without per-dataset tuning, as the thresholds are relative to the normalized mesh (e.g., $0.1$ is about $6\%$ of the object extent); Table~\ref{tab:prop_ablation} reports their sensitivity.

\subsubsection{Matching.}
Both stages use a GPU-parallel auction solver (Bertsekas' algorithm with $\varepsilon$-scaling), which costs $\mathcal{O}(NM\log(C/\varepsilon))$ on the $N\times M$ cost matrix ($N\le30$k foreground vertices, $M\le16$k depth pixels). In each round, all vertices bid in parallel in a single fused kernel pass, with no per-row augmenting-path search, far below the $\mathcal{O}(N^3)$ of the Hungarian algorithm. Empirically, a log-log fit of runtime against $NM$ gives an exponent of $0.889$, and neither $N$ nor $M$ grows with tessellation or dataset size. On GSO, refinement takes $7.5$\,s per instance on average, of which matching accounts for $34.5\%$ and cot-Laplacian propagation for $61.9\%$; propagation, not matching, is the bottleneck.

\begin{table}[t]
\centering
\caption{Comparison with single-view RGBD-to-3D methods.}
\label{tab:rgbd}
\vspace{-2mm}
\scalebox{0.78}{%
\setlength{\tabcolsep}{3pt}
\begin{tabular}{lcccccccc}
\toprule
\multirow{2}{*}{Model} & \multicolumn{4}{c}{GSO} & \multicolumn{4}{c}{OmniObject3D} \\
\cmidrule(lr){2-5} \cmidrule(lr){6-9}
& CD & FS(0.02) & FS(0.05)  & FS(0.1) & CD  & FS(0.02) & FS(0.05) & FS(0.1)  \\
\midrule
\textit{Input point cloud} & 0.0396 & 0.5365 & 0.7350 & 0.8621 & 0.0480 & 0.4298 & 0.6858 & 0.8495 \\
NU-MCC~\cite{lionar2023nu} & 0.0386 & 0.5023 & 0.7415 & 0.8925 & 0.0429 & 0.4567 & 0.7243 & 0.8830 \\
MCC~\cite{wu2023multiview} & 0.0342 & 0.4852 & 0.7694 & 0.9311 & 0.0425 & 0.4124 & 0.7018 & 0.8935 \\
\midrule
Affostruction~\cite{park2026affostruction} & 0.0243 & 0.6698 & 0.8713 & 0.9519 & 0.0251 & 0.6541 & 0.8681 & 0.9485 \\
Affostruction+\textit{ours} (depth only) & 0.0207 & 0.7232 & 0.9036 & 0.9626 & 0.0218 & 0.7003 & 0.8923 & 0.9573 \\
Affostruction+\textit{ours} & 0.0212 & 0.7059 & 0.9003 & 0.9628 & 0.0222 & 0.6900 & 0.8882 & 0.9576 \\
\bottomrule
\end{tabular}%
}

\end{table}
\afterpage{%
\begin{table}[t]
\centering
  \caption{Comparison with REFINE~\cite{leung2022blackREFINE} on its 6-instance demo set.}
  \vspace{-2mm}
  \scalebox{0.78}{%
  \setlength{\tabcolsep}{2pt}
  \begin{tabular}{lcccccc}
  \toprule
  \multirow{2}{*}{Method (CD~$\times 10^3$~$\downarrow$ / Time (s)~$\downarrow$)}
   & \multicolumn{2}{c}{REFINE (1.5k)} & \multicolumn{2}{c}{InstMesh (1.5k)} & \multicolumn{2}{c}{InstMesh (10k)} \\
   \cmidrule(lr){2-3}\cmidrule(lr){4-5}\cmidrule(lr){6-7}
   & CD & Time & CD & Time & CD & Time \\
  \midrule
  Initial mesh                        & 21.58 & --   & 8.31 & --   & 8.30  & --    \\
  REFINE~\cite{leung2022blackREFINE}  & 8.38  & 32.0 & 9.18 & 32.9 & 25.21 & 103.2 \\
  Ours \textit{(w/ pred.\ depth scaled to pred.\ mesh)} & 23.09 & 1.22 & 8.48 & 1.45 & \textbf{8.18} & 7.04  \\
  \bottomrule
  \end{tabular}
  }
  \label{tab:refine_comparison}

\vspace{3mm}
   \caption{Comparison with MeTTA~\cite{kim2024metta} on 9 Pix3D~\cite{sun2018pix3d} instances.}
  \vspace{-2mm}
  \scalebox{0.78}{%
  \setlength{\tabcolsep}{4pt}
  \begin{tabular}{lccccc}
  \toprule
  Method & CD~($\times10^3$)~$\downarrow$ & PSNR~$\uparrow$ & LPIPS~$\downarrow$ & CLIP~$\uparrow$ & Time~(s)~$\downarrow$ \\
  \midrule
  Initial mesh                                           & 18.2 & 13.64 & 0.299 & 0.736 & --     \\
  MeTTA~\cite{kim2024metta}                              & 48.3 & \textbf{14.03} & \textbf{0.278} & 0.650 & $\sim$1.8k \\
  Ours \textit{(w/ GT depth)}                            & \textbf{16.9} & 13.49 & 0.295 & \textbf{0.748} & 1.8 \\
  Ours \textit{(w/ pred.\ depth scaled to pred.\ mesh)}  & 25.3 & 13.41 & 0.303 & 0.649 & 2.6 \\
  \bottomrule
  \end{tabular}
  }
  \label{tab:metta_comparison}

\end{table}}

\subsection{Model-agnostic Single-view Mesh Refinement}
\label{subsec:model_agnostic}
We apply our refinement to five reconstructors with different representations: LGM~\cite{tang2024lgm} decodes multi-view Gaussian features with an asymmetric U-Net, CRM~\cite{wang2024crm} extracts a FlexiCubes mesh from a high-resolution triplane built from orthographic views, InstantMesh~\cite{xu2024instantmesh} pairs a multi-view diffusion prior with a sparse-view LRM, SF3D~\cite{boss2025sf3d} regresses a textured mesh in a single feed-forward pass, and SPAR3D~\cite{huang2025spar3d} samples a point set by point diffusion before image-conditioned meshing. It lowers CD and raises FS at both thresholds for every backbone on both datasets (Table~\ref{tab:table_main}), with the largest gains for weaker initial meshes and at the stricter $0.05$ threshold, which indicates that it corrects fine-grained misalignment rather than only coarse overlap. The rendered images also improve in SSIM, PSNR, and LPIPS everywhere (Table~\ref{tab:table_2d}).

\subsection{Comparison with Single-view RGBD Methods}
In Sec.~\ref{subsec:model_agnostic}, our refinement uses the GT depth, which the RGB-to-3D backbones never see, and the gains may merely reflect this extra input. We therefore compare with methods that use depth from the start, all given the same RGB-D observation with the GT depth and camera pose (Table~\ref{tab:rgbd}), and test whether our refinement still helps a reconstructor that already consumes depth. The RGBD-to-3D completion methods MCC~\cite{wu2023multiview} and NU-MCC~\cite{lionar2023nu} complete the back-projected partial point cloud into a point set, and the \emph{Input point cloud} is the raw back-projection. We evaluate these point sets in the input camera frame, as a global-rotation search on point sets confounds alignment with geometric error. Affostruction~\cite{park2026affostruction} instead reconstructs a full mesh from the same input, and we report its meshes before and after our refinement (+\textit{ours}).
The learned completion methods improve only modestly over the raw input point cloud. Affostruction already outperforms them, and our depth-guided refinement improves it further on both datasets. The symmetry stage brings no additional gain on this backbone, suggesting that a depth-conditioned generator leaves little for mirroring to correct. The refined RGB-to-3D backbones in Table~\ref{tab:table_main} also surpass the completion methods.

\subsection{Comparison with Test-time Refinement Methods}
\subsubsection{Comparison with REFINE.}
REFINE~\cite{leung2022blackREFINE} refines a given mesh at test time by optimizing a small per-instance network under a rendered-silhouette loss and a fixed symmetry prior, without depth. Lacking a shared benchmark, we compare on its 6-instance demo set (Table~\ref{tab:refine_comparison}), starting all methods from the same three initial meshes, REFINE's own template and InstantMesh at 1.5k and 10k vertices, with the camera released with REFINE. Ours uses pseudo-depth affinely fit to the depth rendered from the initial mesh, and we compute CD with REFINE's evaluator. REFINE improves its own coarse template, but its mask-only supervision axis-stretches both InstantMesh meshes and drops them \emph{below} the initial mesh. Our depth matching improves the 10k-vertex mesh, whereas on the two 1.5k-vertex meshes our pseudo-depth refinement falls below the initial mesh, indicating that it needs a sufficient vertex count. Being closed-form in every stage, our method is also an order of magnitude faster.

\subsubsection{Comparison with MeTTA.}
MeTTA~\cite{kim2024metta} refines a given mesh at test time by distilling a 2D diffusion prior over ${\sim}1.5$k iterations, jointly optimizing geometry, texture, and lighting without depth. We compare on its 9 released Pix3D~\cite{sun2018pix3d} instances (Table~\ref{tab:metta_comparison}), starting all methods from the mesh shipped with MeTTA and using the Pix3D camera annotation. Ours uses either the GT depth rendered from the CAD model or pseudo-depth affinely fit to the depth rendered from the initial mesh and warped into the GT silhouette, and we compute the 2D metrics on geometry-only renders at the input view. Optimizing appearance rather than observed geometry lets MeTTA drift far from the initial mesh, whereas ours improves it with GT depth and stays much closer with pseudo-depth, at a runtime orders of magnitude lower. Note that the Pix3D CAD GT differs from the photographed object, which penalizes refinement toward the observed depth. MeTTA scores better on PSNR and LPIPS at the input view, which it directly optimizes; against renders of the GT mesh at novel views, which Table~\ref{tab:novel_view} evaluates, it falls below the initial mesh at every view.

\begin{table}[t]
\centering
\caption{Ablation of depth and symmetry guidance (CD).}
\label{tab:ablation}
\vspace{-2mm}
\resizebox{\textwidth}{!}{%
\begin{tabular}{lcccccccccc}
\toprule
\multirow{2}{*}{Model} & \multicolumn{5}{c}{GSO} & \multicolumn{5}{c}{OmniObject3D} \\
\cmidrule(lr){2-6} \cmidrule(lr){7-11}
 & LGM & CRM & SF3D & SPAR3D & InstMesh & LGM & CRM & SF3D & SPAR3D & InstMesh  \\
\midrule
Initial mesh (RGB$\rightarrow$3D)
& 0.0438 & 0.0365 & 0.0352 & 0.0356 & 0.0283
& 0.0397 & 0.0334 & 0.0311 & 0.0331 & 0.0320 \\
\textit{w/ pseudo-depth (scaled to GT)}
& 0.0367 & 0.0315 & 0.0313 & 0.0320 & 0.0272
& 0.0344 & 0.0296 & 0.0286 & 0.0309 & 0.0296 \\
\textit{w/ depth}
& 0.0323 & 0.0253 & 0.0244 & 0.0253 & 0.0212
& 0.0302 & 0.0243 & 0.0217 & 0.0241 & 0.0236 \\
\textit{w/ depth, symmetry}
& 0.0306 & 0.0248 & 0.0236 & 0.0248 & 0.0205
& 0.0291 & 0.0233 & 0.0210 & 0.0238 & 0.0229 \\
\bottomrule
\end{tabular}%
}

\vspace{3mm}
\caption{Refinement on symmetric and asymmetric objects (CD).}
\label{tab:asymmetric}
\vspace{-2mm}
\resizebox{\textwidth}{!}{%
\setlength{\tabcolsep}{2pt}
\begin{tabular}{lcccccccccc}
\toprule
\multirow{2}{*}{Model} & \multicolumn{5}{c}{GSO \textit{Symmetric}} & \multicolumn{5}{c}{OmniObject3D \textit{Symmetric}} \\
\cmidrule(lr){2-6} \cmidrule(lr){7-11}
 & LGM & CRM & InstMesh & SF3D & SPAR3D & LGM & CRM & InstMesh & SF3D & SPAR3D \\
 \midrule
Initial mesh (RGB$\rightarrow$3D)  & 0.0419 & 0.0363 & 0.0247 & 0.0354 & 0.0363 & 0.0342 & 0.0262 & 0.0297 & 0.0235 & 0.0257 \\
+ Depth & 0.0306 & 0.0265 & 0.0185 & 0.0244 & 0.0262 & 0.0263 & 0.0201 & 0.0232 & 0.0172 & 0.0184 \\
+ Depth, Symmetry & 0.0290 & 0.0251 & 0.0174 & 0.0224 & 0.0247 & 0.0252 & 0.0191 & 0.0217 & 0.0168 & 0.0177 \\
\midrule
\multirow{2}{*}{Model} & \multicolumn{5}{c}{GSO \textit{Asymmetric}} & \multicolumn{5}{c}{OmniObject3D \textit{Asymmetric}} \\
\cmidrule(lr){2-6} \cmidrule(lr){7-11}
 & LGM & CRM & InstMesh & SF3D & SPAR3D & LGM & CRM & InstMesh & SF3D & SPAR3D \\
 \midrule
Initial mesh (RGB$\rightarrow$3D)  & 0.0462 & 0.0369 & 0.0327 & 0.0350 & 0.0346 & 0.0399 & 0.0337 & 0.0321 & 0.0314 & 0.0333 \\
+ Depth & 0.0343 & 0.0238 & 0.0246 & 0.0244 & 0.0242 & 0.0304 & 0.0245 & 0.0237 & 0.0218 & 0.0243 \\
+ Depth, Symmetry & 0.0326 & 0.0245 & 0.0245 & 0.0251 & 0.0250 & 0.0292 & 0.0235 & 0.0230 & 0.0211 & 0.0241 \\
\bottomrule
\end{tabular}
}

\end{table}

\subsection{Ablation Study}
\subsubsection{Depth guidance.}
Table~\ref{tab:ablation} adds the components one at a time on all five backbones. Depth guidance alone, which anchors the visible vertices to the depth map, gives the largest reduction in CD on both datasets; with pseudo-depth rescaled to the GT depth range, the gain is smaller but consistent.

\subsubsection{Symmetry guidance.}
Adding the symmetry stage, which mirrors the depth-guided corrections into occluded regions, lowers CD further for every backbone (Table~\ref{tab:ablation}). To see where symmetry helps, Table~\ref{tab:asymmetric} splits the objects into symmetric and asymmetric sets by Reflect3D~\cite{li2025symmetry0strikes} labels: the symmetry stage gives its largest gains on symmetric objects and can slightly hurt asymmetric ones, whose mirrored corrections do not match the true geometry.

\subsubsection{Propagation and matching thresholds.}
Table~\ref{tab:prop_ablation} examines the propagation and the matching thresholds on InstantMesh/GSO. To isolate the propagation, we compare three outputs: \emph{handles only}, which moves the matched vertices and leaves the rest unchanged; \emph{+ propagation}, which spreads these displacements with the smoothness solver; and \emph{+ symmetry}, which adds the reflected handles. At our threshold of $0.1$, moving only the matched vertices already recovers most of the gain over the initial mesh (Table~\ref{tab:table_main}), the propagation further lowers CD and raises the strict FS(0.02), and symmetry lowers CD further. To test the sensitivity to the thresholds, we vary the depth-matching threshold from $0.02$ to $0.5$ and the symmetry-matching threshold from $0.1$ to $0.4$; \#matched reports the mean number of depth-matched vertices out of about $10.4$k visible ones. The result is stable between $0.1$ and $0.2$. A tighter threshold matches fewer vertices and weakens the correction, whereas a much looser one admits wrong correspondences and falls below the initial mesh. The symmetry-matching threshold barely changes the result.

\begin{table}[t]
\centering
\caption{Propagation ablation and matching-threshold sweep (InstantMesh, GSO).}
\label{tab:prop_ablation}
\vspace{-2mm}
\resizebox{\textwidth}{!}{%
\setlength{\tabcolsep}{4pt}
\begin{tabular}{lcccccccccc}
\toprule
\multirow{2}{*}{Threshold} & \multirow{2}{*}{\#matched}
 & \multicolumn{3}{c}{handles only} & \multicolumn{3}{c}{+ propagation} & \multicolumn{3}{c}{+ symmetry} \\
\cmidrule(lr){3-5}\cmidrule(lr){6-8}\cmidrule(lr){9-11}
 & & CD & FS(0.02) & FS(0.05) & CD & FS(0.02) & FS(0.05) & CD & FS(0.02) & FS(0.05) \\
\midrule
\multicolumn{11}{l}{\textit{Depth-matching threshold}}\\
\quad 0.02          & 1583 & 0.0269 & 0.585 & 0.843 & 0.0262 & 0.602 & 0.843 & 0.0258 & 0.609 & 0.847 \\
\quad 0.05          & 2593 & 0.0241 & 0.652 & 0.866 & 0.0233 & 0.666 & 0.863 & 0.0227 & 0.674 & 0.871 \\
\quad 0.1 (ours)    & 3093 & 0.0221 & 0.679 & 0.890 & 0.0212 & 0.701 & 0.886 & 0.0205 & 0.707 & 0.895 \\
\quad 0.2           & 3362 & 0.0215 & 0.683 & 0.894 & 0.0210 & 0.704 & 0.890 & 0.0203 & 0.708 & 0.898 \\
\quad 0.5           & 3499 & 0.0384 & 0.606 & 0.810 & 0.0289 & 0.658 & 0.844 & 0.0310 & 0.641 & 0.832 \\
\midrule
\multicolumn{11}{l}{\textit{Symmetry-matching threshold (depth threshold $0.1$)}}\\
\quad 0.1           & 3093 & \multicolumn{3}{c}{--} & \multicolumn{3}{c}{--} & 0.0205 & 0.707 & 0.895 \\
\quad 0.2 (ours)    & 3093 & \multicolumn{3}{c}{--} & \multicolumn{3}{c}{--} & 0.0205 & 0.707 & 0.895 \\
\quad 0.4           & 3093 & \multicolumn{3}{c}{--} & \multicolumn{3}{c}{--} & 0.0206 & 0.707 & 0.895 \\
\bottomrule
\end{tabular}%
}

\end{table}

\begin{table}[t]
\centering
  \caption{Robustness to depth source and camera pose (InstantMesh, GSO).}
  \vspace{-2mm}
  \scalebox{0.73}{%
  \setlength{\tabcolsep}{4pt}
  \begin{tabular}{lccc}
  \toprule
  Method & CD~$\downarrow$ & FS(0.02)~$\uparrow$ & FS(0.05)~$\uparrow$ \\
  \midrule
  Initial mesh & 0.0283 & 0.5581 & 0.8358 \\
  \midrule
  \multicolumn{4}{l}{\textit{Depth source (GT pose, depth-only)}}\\
  \quad w/ GT depth                                   & 0.0212 & 0.7008 & 0.8856 \\
  \quad w/ pseudo-depth (scaled to GT)                & 0.0272 & 0.5597 & 0.8509 \\
  \quad w/ pseudo-depth (no GT, scaled to pred.\ mesh)& 0.0275 & 0.5604 & 0.8494 \\
  \midrule
  \multicolumn{4}{l}{\textit{Camera pose (GT depth + symmetry)}}\\
  \quad w/ GT pose                                    & 0.0205 & 0.7070 & 0.8950 \\
  \quad w/ est.\ pose (FoundationPose~\cite{wen2024foundationpose}) & 0.0262 & 0.6006 & 0.8557 \\
  \midrule
  \multicolumn{4}{l}{\textit{Both estimated (pseudo-depth scaled to GT + est.\ pose, with symmetry)}}\\
  \quad w/ pseudo-depth and est.\ pose & 0.0298 & 0.5004 & 0.8302 \\
  \bottomrule
  \end{tabular}
  }
  \label{tab:robust}

\vspace{3mm}
  \caption{Real-world results on YCB-V~\cite{xiang2018posecnn} (CD~$\downarrow$ / FS(0.05)~$\uparrow$).}
  \vspace{-2mm}
  \resizebox{\textwidth}{!}{%
  \setlength{\tabcolsep}{4pt}
  \begin{tabular}{lccccc}
  \toprule
  \multirow{2}{*}{Method}
   & \multicolumn{2}{c}{symmetry class} & \multicolumn{3}{c}{visibility} \\
   \cmidrule(lr){2-3}\cmidrule(lr){4-6}
   & \textit{symmetric} ($N{=}103$) & \textit{asymmetric} ($N{=}298$) & $[0.5,0.7)$ ($N{=}39$) & $[0.7,0.9)$ ($N{=}78$) & $[0.9,1]$ ($N{=}284$) \\
  \midrule
  Initial mesh              & 0.0530 / 0.634 & 0.0417 / 0.725 & 0.0685 / 0.500 & 0.0619 / 0.552 & 0.0365 / 0.771 \\
  Refined \textit{(ours)}   & \textbf{0.0503} / \textbf{0.650} & 0.0416 / 0.720 & \textbf{0.0645} / \textbf{0.519} & \textbf{0.0579} / \textbf{0.584} & 0.0372 / 0.760 \\
  \bottomrule
  \end{tabular}
  }
  \label{tab:ycbv_symmetry}

\end{table}

\subsection{Analysis}
\subsubsection{Robustness to camera pose and depth.}
In practice, the depth and the camera pose come from sensors or estimators rather than the GT. Table~\ref{tab:robust} tests three such settings on InstantMesh/GSO. \emph{Estimated depth:} we replace the GT depth with UniDepth-V2~\cite{piccinelli2025unidepthv2} pseudo-depth, either rescaled to the GT depth range or affinely fit to the depth rendered from the initial mesh (no GT), keeping the GT pose and refining without symmetry. Both variants still improve over the initial mesh, and the no-GT fit performs on par with the GT-scaled one. \emph{Estimated pose:} we replace the GT pose with a FoundationPose~\cite{wen2024foundationpose} estimate computed from the RGB image, the no-GT pseudo-depth, the instance mask, and the initial mesh, keeping the GT depth and the symmetry stage. The refinement still reduces CD and improves the F-score at every threshold. \emph{Both estimated:} combining the estimated pose with the GT-scaled pseudo-depth drops the result below the initial mesh; the refinement thus needs at least one reliable input.

\begin{figure*}[t!]
    \centering
    \includegraphics[width=\textwidth]{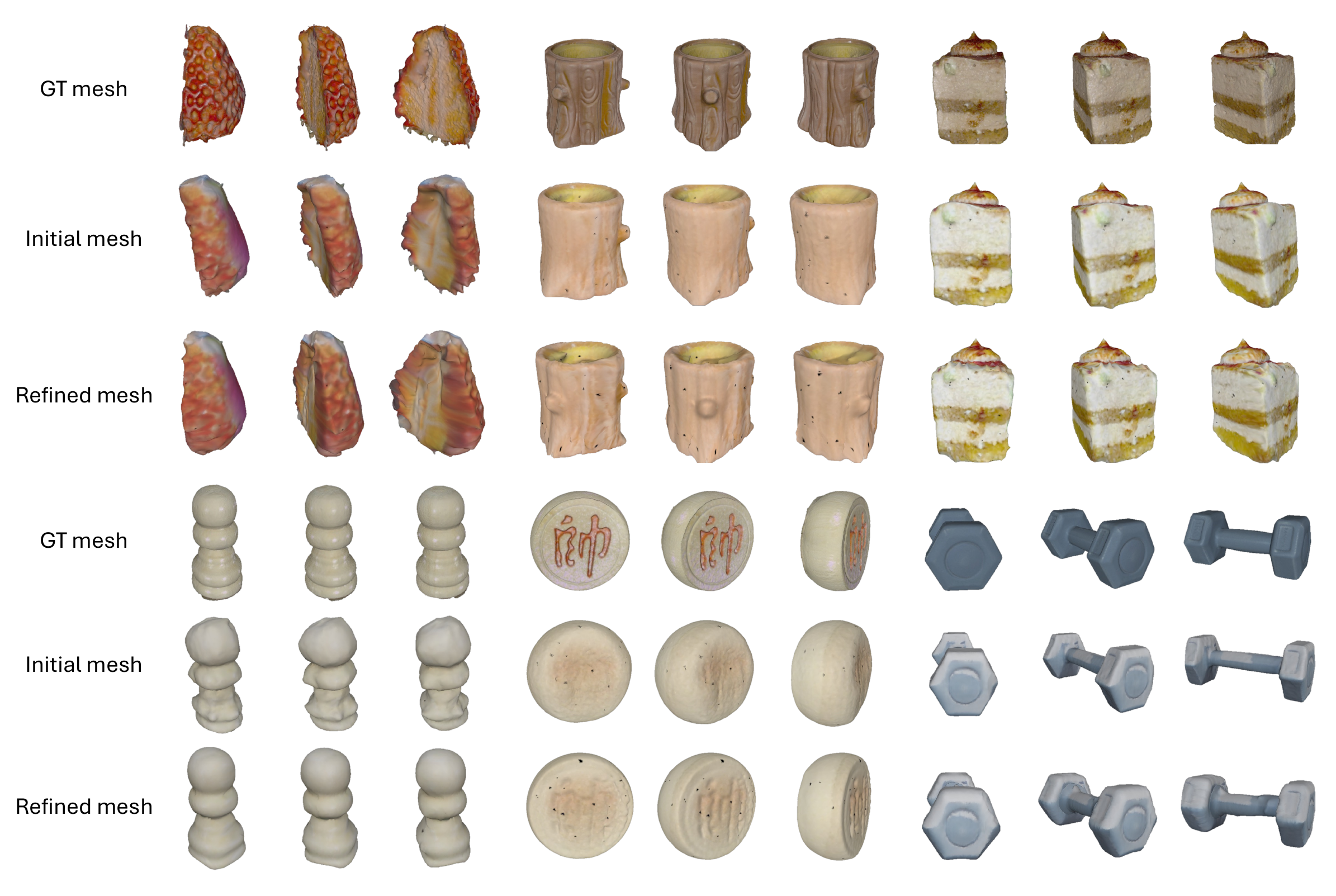}
    \caption{Qualitative results on OmniObject3D. For each mesh, we render three views: the center view matches the input view, and the left/right views are novel viewpoints.}
    \label{fig:qual_omni}
    \vspace{-2mm}
\end{figure*}

\subsubsection{Real-world generalization.}
Our main results use synthetic renders, whereas real captures bring noisy sensor depth. We therefore evaluate on YCB-V~\cite{xiang2018posecnn}: we take every tenth test keyframe ($401$ instances), run InstantMesh on the object crop, and refine with the annotated camera pose and the sensor depth, applying a guarded ICP at evaluation. Table~\ref{tab:ycbv_symmetry} groups the instances by BOP symmetry class and by the visibility of the object in the input frame. Across symmetry classes, the \emph{same} pipeline improves symmetric objects and leaves asymmetric ones nearly unchanged, matching the synthetic trend: depth drives the gain, and symmetry helps where present. Across visibility levels, refinement gives a clear gain on partially visible objects, whereas almost fully visible objects are already accurate and change only marginally. As most YCB-V instances are almost fully visible, the average gain is smaller than on synthetic data.

\begin{figure*}[t!]
    \centering
    \includegraphics[width=\textwidth]{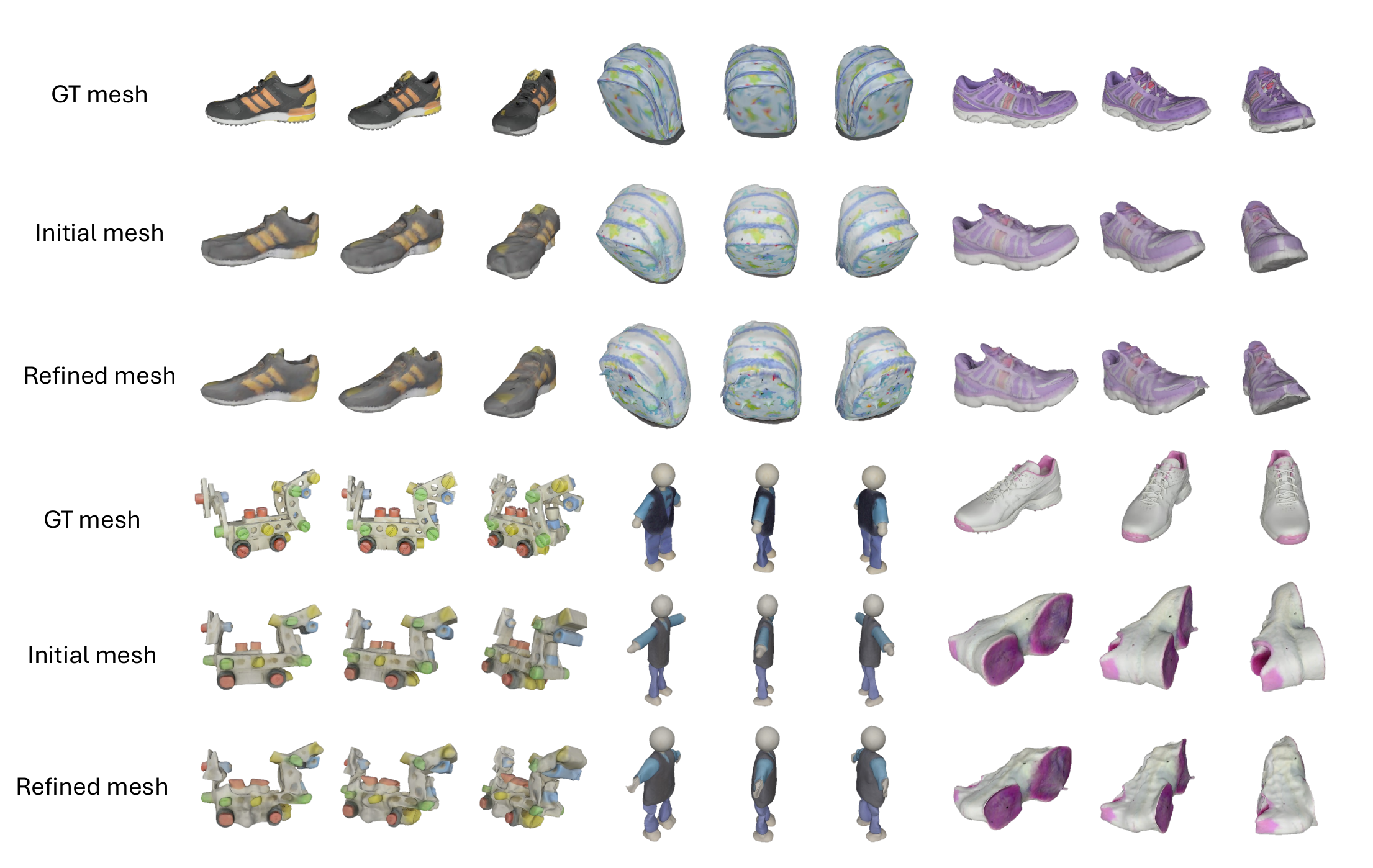}
    \caption{Qualitative results on GSO. For each mesh, we render three views: the center view matches the input view, and the left/right views are novel viewpoints.}
    \label{fig:qual_gso}
    \vspace{-2mm}
\end{figure*}

\subsection{Qualitative Results}
Across object categories (Figs.~\ref{fig:qual_omni} and~\ref{fig:qual_gso}), the refinement improves source-view consistency, with cleaner silhouettes, flatter planar regions, and more plausible surface relief on the visible side. The improvements are most noticeable when the initial mesh is roughly aligned with the input view, and the initial prediction bounds the result. The second row of Fig.~\ref{fig:qual_gso} shows failure cases: thin or dense structures can be distorted, and imperfect alignment biases the occluded regions.

\section{Conclusion}
\label{sec:conclusion}

We presented a lightweight, plug-and-play RGBD-to-3D refinement framework for any RGB-to-3D reconstructor, requiring no retraining. It corrects the observed surface by depth matching, mirrors these corrections onto the occluded side by symmetry, and propagates them with a smoothness solver, each in closed form. Across five backbones and two datasets, it yields consistent gains that persist with pseudo-depth, larger symmetry gains on symmetric objects, and favorable accuracy and runtime against prior refinement. Future work includes learned completion for severe occlusions and symmetry-plane confidence.
\smallbreak
\noindent \textbf{Acknowledgements.}
This work was supported by Samsung Electronics Co., Ltd.  (IO240508-09825-01) and IITP grants (RS-2024-00457882: National AI Research Lab Project (50\%), RS-2022-II220959: Few-Shot Learning of Causal Inference in Vision and Language for Decision Making (30\%), RS-2022-II220290: Visual Intelligence for Space-Time Understanding \& Generation (20\%) ) funded by Ministry of Science and ICT, Korea.
\looseness=-1

\bibliographystyle{splncs04}
\bibliography{main}

\clearpage
\appendix
\begin{center}
{\Large\bfseries Supplementary Material}
\end{center}
\vspace{0.5em}

\section{Sensitivity Analysis}

\subsubsection{Camera pose and intrinsics.}
\begin{table}[h]
\vspace{-4mm}
\centering
\caption{CD improvement ($\times 10^{-3}$) under camera perturbations (InstantMesh).}
\vspace{-2mm}
\label{tab:table_sweep}
\resizebox{\textwidth}{!}{%
\setlength{\tabcolsep}{3pt}
\begin{tabular}{llccccccccccccc}
\toprule
& \multirow{2}{*}{$\Delta$ CD $\downarrow$} & \multirow{2}{*}{baseline} & \multicolumn{4}{c}{camera pose (polar)} & \multicolumn{4}{c}{camera pose (azimuth)} & \multicolumn{4}{c}{intrinsics (focal)} \\
\cmidrule(lr){4-7}\cmidrule(lr){8-11}\cmidrule(lr){12-15}
& & & $\pm1^\circ$ & $\pm2^\circ$ & $\pm5^\circ$ & $\pm10^\circ$ & $\pm1^\circ$ & $\pm2^\circ$ & $\pm5^\circ$ & $\pm10^\circ$ & \scalebox{0.85}{$+5\%$} & \scalebox{0.85}{$+10\%$} & \scalebox{0.85}{$+15\%$}& \scalebox{0.85}{$+20\%$} \\
\midrule
\multirow{2}{*}{\rotatebox{90}{GSO}}
& \textit{w/ depth} & -7.19 & -7.03 & -6.65 & -5.14 & -2.04 & -7.08 & -6.79 & -5.68 & -3.42 & -6.48 & -5.01 & -3.48 & -2.00 \\
& \textit{w/ depth, sym.} & -7.88 & -7.63 & -7.26 & -5.68 & -2.53 & -7.76 & -7.55 & -6.52 & -4.26 & -7.33 & -5.87 & -4.25 & -2.67 \\
\midrule
\multirow{2}{*}{\rotatebox{90}{Omni}}
& \textit{w/ depth} & -8.34 & -8.20 & -7.83 & -6.44 & -3.67 & -8.27 & -8.10 & -7.32 & -5.76 & -7.34 & -5.41 & -3.40 & -1.51 \\
& \textit{w/ depth, sym.} & -9.11 & -9.01 & -8.65 & -7.13 & -4.13 & -9.03 & -8.94 & -8.23 & -6.67 & -8.28 & -6.32 & -4.25 & -2.23 \\
\bottomrule
\end{tabular}%
}
\end{table}

We perturb only the input-view camera on every fifth object of GSO and OmniObject3D (InstantMesh, GT depth): polar or azimuth rotations of $\pm\{1,2,5,10\}^\circ$ and focal-length inflation of $\{5,10,15,20\}\%$. Table~\ref{tab:table_sweep} reports the change in CD relative to the initial mesh (more negative is better). Refinement stays beneficial under every perturbation and degrades gracefully (on GSO, from $-7.19$ to $-2.04$ at $\pm10^\circ$ polar rotation and $-2.00$ at $+20\%$ focal inflation), and symmetry-guided refinement is best at every level.

\subsubsection{Depth perturbation.}
On GSO with InstantMesh and the GT pose (initial CD $0.0283$), we perturb the GT depth before back-projection with Gaussian noise ($\sigma$ in \% of the foreground depth range), Gaussian blur (kernel size in pixels), random hole dropout (\% of foreground pixels), or a global scale error (Table~\ref{tab:depth_perturb}). Noise up to $5\%$, blur, and up to $20\%$ missing pixels change CD by at most $0.002$. Only a global scale error is harmful: $2\%$ costs $0.002$--$0.004$ CD, and at $5\%$ the result falls below the initial mesh. The affine fit to the initial mesh removes exactly this error, at a cost of $0.0003$ CD (Table~\ref{tab:robust}).

\begin{table}[!htbp]
\centering
\caption{Depth-perturbation sweep (InstantMesh, GSO; CD).}
\vspace{-2mm}
\label{tab:depth_perturb}
\resizebox{\textwidth}{!}{%
\setlength{\tabcolsep}{4pt}
\begin{tabular}{lcccccccc}
\toprule
\multirow{2}{*}{CD $\downarrow$} & \multirow{2}{*}{unperturbed} & \multicolumn{4}{c}{noise $\sigma$} & \multicolumn{3}{c}{blur} \\
\cmidrule(lr){3-6}\cmidrule(lr){7-9}
& & $0.5\%$ & $1\%$ & $2\%$ & $5\%$ & $3$\,px & $5$\,px & $9$\,px \\
\midrule
\textit{w/ depth}       & 0.0212 & 0.0212 & 0.0212 & 0.0212 & 0.0213 & 0.0213 & 0.0213 & 0.0214 \\
\textit{w/ depth, sym.} & 0.0205 & 0.0205 & 0.0205 & 0.0205 & 0.0205 & 0.0206 & 0.0206 & 0.0207 \\
\bottomrule
\end{tabular}%
}

\vspace{2mm}
\resizebox{\textwidth}{!}{%
\setlength{\tabcolsep}{3pt}
\begin{tabular}{lcccccccccccc}
\toprule
\multirow{2}{*}{CD $\downarrow$} & \multicolumn{5}{c}{holes} & \multicolumn{7}{c}{scale error} \\
\cmidrule(lr){2-6}\cmidrule(lr){7-13}
& $1\%$ & $2\%$ & $5\%$ & $10\%$ & $20\%$ & $+1\%$ & $+2\%$ & $-2\%$ & $+5\%$ & $-5\%$ & $+10\%$ & $-10\%$ \\
\midrule
\textit{w/ depth}       & 0.0215 & 0.0216 & 0.0219 & 0.0223 & 0.0228 & 0.0218 & 0.0232 & 0.0251 & 0.0301 & 0.0325 & 0.0411 & 0.0427 \\
\textit{w/ depth, sym.} & 0.0209 & 0.0211 & 0.0214 & 0.0218 & 0.0224 & 0.0210 & 0.0224 & 0.0248 & 0.0291 & 0.0332 & 0.0391 & 0.0442 \\
\bottomrule
\end{tabular}%
}
\end{table}

\subsubsection{Depth-to-vertex density.}
On GSO with InstantMesh, the GT pose, and the GT depth, we vary the depth back-projection resolution and the vertex cap, which changes the number of depth points by $16\times$ and of visible vertices by $2.7\times$ (Table~\ref{tab:density}; counts are means over the $1{,}030$ objects, and \emph{handles} moves the matched vertices without propagation). CD stays within $0.0203$--$0.0212$. With the sparsest depth, the matching covers only $8\%$ of the visible vertices, yet CD still drops by $25\%$: sparse depth yields fewer handles, not wrong displacements.

\begin{table}[!htbp]
\centering
\caption{Depth-resolution and vertex-density sweep (InstantMesh, GSO; CD).}
\vspace{-2mm}
\label{tab:density}
\small
\setlength{\tabcolsep}{5pt}
\begin{tabular}{llcccccc}
\toprule
\multirow{2}{*}{vertices} & \multirow{2}{*}{depth} & \multicolumn{3}{c}{count} & \multicolumn{3}{c}{CD $\downarrow$} \\
\cmidrule(lr){3-5}\cmidrule(lr){6-8}
 & & visible & depth & matched & handles & depth & depth, sym. \\
\midrule
\multirow{3}{*}{30k} & $64^2$         & 10393 & 892   & 805  & 0.0241 & 0.0221 & 0.0212 \\
                     & $128^2$ (ours) & 10393 & 3568  & 3093 & 0.0221 & 0.0212 & 0.0205 \\
                     & $256^2$        & 10393 & 14271 & 8036 & 0.0211 & 0.0209 & 0.0204 \\
\midrule
\multirow{3}{*}{10k} & $64^2$         & 3902  & 892   & 788  & 0.0234 & 0.0220 & 0.0212 \\
                     & $128^2$        & 3902  & 3568  & 2335 & 0.0219 & 0.0218 & 0.0210 \\
                     & $256^2$        & 3902  & 14271 & 3487 & 0.0219 & 0.0205 & 0.0203 \\
\bottomrule
\end{tabular}
\end{table}

\section{Symmetry Analysis}

\begin{figure*}[h]
    \centering
    \includegraphics[width=\textwidth]{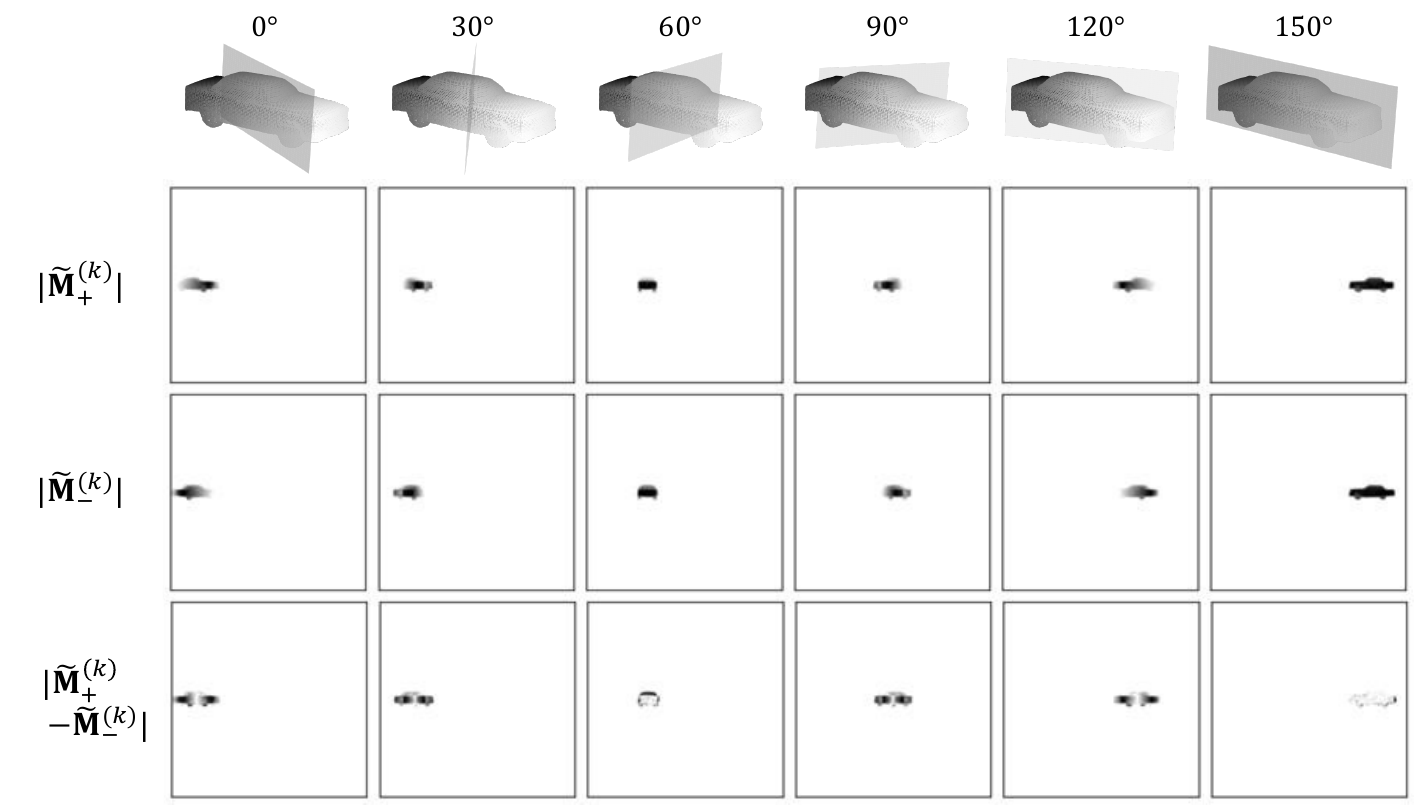}
    \caption{Rasterized signed distance maps for different candidate symmetry planes.}
    \label{fig:sd_map}
\end{figure*}

\subsubsection{Symmetry plane detection.}
Figure~\ref{fig:sd_map} visualizes the two signed-distance maps of the plane-detection criterion (Sec.~\ref{subsec:symmetry}) for several candidate planes. The score varies smoothly across the swept planes and has a single clear minimum at the correct plane, which keeps the selection stable.

\subsubsection{Symmetry-plane source.}
We replace our plane with Reflect3D's~\cite{li2025symmetry0strikes} learned prediction on the $1{,}023$ GSO objects for which Reflect3D returns a valid plane (InstantMesh, GT pose and depth). This hurts refinement (Table~\ref{tab:sym_ablation}): for objects with several symmetries its top-1 may be a top or front plane, whereas our camera-frame search returns the left/right plane that mirrors the visible side onto the occluded one, at ${\sim}0.27$\,s per object with no extra network.

\begin{table}[h]
\centering
  \caption{Symmetry-plane source ablation (InstantMesh, GSO).}
  \label{tab:sym_ablation}
  \vspace{-2mm}
  \resizebox{0.82\textwidth}{!}{%
  \setlength{\tabcolsep}{2pt}
  \begin{tabular}{lcccc}
  \toprule
  \multirow{2}{*}{Method}
   & \multicolumn{2}{c}{\textit{symmetric}} & \multicolumn{2}{c}{\textit{asymmetric}} \\
   \cmidrule(lr){2-3}\cmidrule(lr){4-5}
   & CD~$\downarrow$ & FS(0.05)~$\uparrow$ & CD~$\downarrow$ & FS(0.05)~$\uparrow$ \\
  \midrule
  Initial mesh                              & 0.0247 & 0.873 & 0.0329 & 0.787 \\
  \textit{w/ GT depth}                      & 0.0185 & 0.912 & 0.0247 & 0.851 \\
  \textit{w/ depth + symmetry (ours)}       & \textbf{0.0174} & \textbf{0.924} & \textbf{0.0246} & \textbf{0.857} \\
  \textit{w/ depth + symmetry (Reflect3D~\cite{li2025symmetry0strikes})} & 0.0217 & 0.883 & 0.0284 & 0.818 \\
  \bottomrule
  \end{tabular}
  }

\end{table}

\subsubsection{Failure analysis.}
For each GSO object, we measure the change from the symmetry stage, $\Delta=\mathrm{CD}(\text{depth+sym.})-\mathrm{CD}(\text{depth})$ (negative is better), and count the object as degraded when $\Delta>0.001$. Table~\ref{tab:sym_failure} splits the objects by Reflect3D label and also reports the thin-structure decile (top $10\%$ of surface area over bounding-box volume). The mean effect is a gain on symmetric objects and at most $+0.0008$ on asymmetric ones, but $16$--$33\%$ of the symmetric and $21$--$48\%$ of the asymmetric objects degrade ($24$--$32\%$ of all objects on OmniObject3D). For InstantMesh, $90$ of the $254$ degraded objects are symmetric-labelled, where a wrong plane is the likely cause, and $164$ are asymmetric. The thin-structure decile behaves like the full set. The score margin of the plane search does not predict $\Delta$ (Spearman $|\rho|\le0.13$), gating on it never beats always-on symmetry, and even an oracle per-object choice gains only $0.0007$ CD. We leave confidence estimation to future work.

\begin{table}[!htbp]
\centering
\caption{Per-object effect of the symmetry stage on GSO.}
\vspace{-2mm}
\label{tab:sym_failure}
\resizebox{\textwidth}{!}{%
\setlength{\tabcolsep}{3pt}
\begin{tabular}{lcccccccccc}
\toprule
\multirow{2}{*}{Backbone}
 & \multicolumn{4}{c}{\textit{symmetric} ($N{=}572$)} & \multicolumn{4}{c}{\textit{asymmetric} ($N{=}458$)} & \multicolumn{2}{c}{\textit{thin} ($N{=}103$)} \\
\cmidrule(lr){2-5}\cmidrule(lr){6-9}\cmidrule(lr){10-11}
 & mean $\Delta$ & \% degr. & p90 & max & mean $\Delta$ & \% degr. & p90 & max & mean $\Delta$ & \% degr. \\
\midrule
LGM         & $-0.0016$ & 22 & $+0.0023$ & $+0.0113$ & $-0.0017$ & 21 & $+0.0032$ & $+0.0112$ & $-0.0023$ & 17 \\
CRM         & $-0.0014$ & 33 & $+0.0039$ & $+0.0149$ & $+0.0006$ & 41 & $+0.0043$ & $+0.0181$ & $-0.0002$ & 41 \\
SF3D        & $-0.0020$ & 21 & $+0.0025$ & $+0.0084$ & $+0.0007$ & 47 & $+0.0046$ & $+0.0124$ & $-0.0009$ & 20 \\
SPAR3D      & $-0.0015$ & 24 & $+0.0029$ & $+0.0107$ & $+0.0008$ & 48 & $+0.0048$ & $+0.0132$ & $+0.0007$ & 36 \\
InstantMesh & $-0.0012$ & 16 & $+0.0016$ & $+0.0093$ & $-0.0001$ & 36 & $+0.0032$ & $+0.0106$ & $-0.0003$ & 17 \\
\bottomrule
\end{tabular}%
}
\end{table}

\section{Implementation Details}
\label{sec:supp_impl}

\subsubsection{Rendering and views.}
The 12 azimuths of the 24 rendered views are spaced by $15^\circ$, from $-82.5^\circ$ to $82.5^\circ$.

\subsubsection{Depth- and symmetry-guided stages.}
We resize the source-view depth to $128\times128$ with bilinear interpolation. The auction solver uses $\varepsilon=10^{-3}$ of the cost scale and up to 12 $\varepsilon$-scaling rounds in both stages. Plane detection scores $1{,}260$ candidates ($180$ azimuths $\times$ $7$ elevations), each rasterized into two $512\times512$ maps.

\subsubsection{Propagation stage.}
The boundary-anchor weight is $\gamma=10^{-3}$ and the Tikhonov weight $\tau=10^{-6}$ of the mean diagonal of $L_{ff}$, and Taubin smoothing uses $\lambda=0.5$ and $\nu=-0.5$.

\section{Evaluation Details}

\subsubsection{Protocol.}
The rotation search is coarse-to-fine over azimuth, elevation, and roll, without ICP: we score a $24^3$ grid on a quarter of the sampled points and then a $48^3$ grid within the 64 best coarse cells. We decimate meshes with the quadric error metric and score the initial mesh before decimation. SSIM uses an $11\times11$ Gaussian window ($\sigma=1.5$) and LPIPS an AlexNet backbone.

\subsubsection{Test-time refinement baselines.}
REFINE~\cite{leung2022blackREFINE} runs 400 optimization iterations per instance, and we compute CD with its evaluator after per-candidate ICP. MeTTA~\cite{kim2024metta} distills Zero-1-to-3~\cite{liu2023zero} onto a DMTet~\cite{shen2021deep} mesh, which takes ${\sim}30$\,min per instance on an RTX~3090; as its meshes live in a per-instance frame, we also compute CD after per-candidate ICP. For the pseudo-depth row of Table~\ref{tab:metta_comparison}, we fit UniDepth-V2~\cite{piccinelli2025unidepthv2} to the depth rendered from the initial mesh by min-max matching over overlapping pixels and warp it into the silhouette of the GT model with a thin-plate spline.

\subsubsection{Novel-view evaluation against MeTTA.}
Real images exist only at the input view, so we render the GT CAD mesh as the reference at the input view and at 7 novel azimuths and render every mesh geometry-only (Table~\ref{tab:novel_view}). Against these renders, MeTTA falls below the initial mesh at every view, including the input view, where it scores better against the photo (Table~\ref{tab:metta_comparison}). Ours with GT depth improves the initial mesh at the input view and matches or improves it at novel views, except for CLIP; with pseudo-depth it matches at the input view but falls below at novel views, consistent with its CD in Table~\ref{tab:metta_comparison}.

\begin{table}[!htbp]
\centering
\caption{Evaluation against GT renders at input and novel views (Pix3D).}
\vspace{-2mm}
\label{tab:novel_view}
\resizebox{\textwidth}{!}{%
\setlength{\tabcolsep}{4pt}
\begin{tabular}{lcccccccc}
\toprule
\multirow{2}{*}{Method} & \multicolumn{2}{c}{PSNR $\uparrow$} & \multicolumn{2}{c}{LPIPS $\downarrow$} & \multicolumn{2}{c}{IoU $\uparrow$} & \multicolumn{2}{c}{CLIP $\uparrow$} \\
\cmidrule(lr){2-3}\cmidrule(lr){4-5}\cmidrule(lr){6-7}\cmidrule(lr){8-9}
 & input & novel & input & novel & input & novel & input & novel \\
\midrule
Initial mesh                              & 19.2 & 19.4 & 0.161 & 0.139 & 0.82 & 0.81 & 0.864 & 0.850 \\
MeTTA~\cite{kim2024metta}                 & 16.4 & 16.6 & 0.282 & 0.277 & 0.68 & 0.62 & 0.774 & 0.780 \\
Ours \textit{(w/ GT depth)}               & \textbf{20.5} & \textbf{19.9} & \textbf{0.127} & \textbf{0.131} & \textbf{0.86} & \textbf{0.81} & \textbf{0.884} & 0.830 \\
Ours \textit{(w/ pred.\ depth scaled to pred.\ mesh)} & 19.2 & 17.7 & 0.156 & 0.194 & 0.83 & 0.67 & 0.794 & 0.794 \\
\bottomrule
\end{tabular}%
}
\end{table}

\subsubsection{RGB-D baselines.}
We run MCC~\cite{wu2023multiview}, NU-MCC~\cite{lionar2023nu}, and Affostruction~\cite{park2026affostruction} from their released weights; Affostruction uses its single-view pipeline. As its symmetry stage brings no gain (Table~\ref{tab:rgbd}), we recommend depth-only refinement for backbones whose occluded side is already conditioned on depth.

\subsubsection{Robustness to depth and pose.}
For Table~\ref{tab:robust}, the no-GT pseudo-depth is fit to the depth rendered from the initial mesh over overlapping pixels, and FoundationPose~\cite{wen2024foundationpose} runs with its default settings.

\subsubsection{YCB-V.}
Beyond the visibility split of Table~\ref{tab:ycbv_symmetry}, $12$ of the $21$ objects and $54\%$ of the instances improve. The rigid residual of the alignment to the CAD model and the initial CD, which is $1.6\times$ that on GSO, limit the gain.

\end{document}